\documentclass[10pt,twocolumn,letterpaper]{article}

\usepackage{cvpr}              % To produce the CAMERA-READY version
\usepackage{adjustbox}
\usepackage{algorithm}
\usepackage[noend]{algpseudocode}
\usepackage[dvipsnames]{xcolor}

\definecolor{cvprblue}{rgb}{0.21,0.49,0.74}
\usepackage[pagebackref,breaklinks,colorlinks,citecolor=cvprblue]{hyperref}

\def\paperID{11586} % *** Enter the Paper ID here
\def\confName{CVPR}
\def\confYear{2024}

\title{MRFP: Learning Generalizable Semantic Segmentation from Sim-2-Real with Multi-Resolution Feature Perturbation}

\author{  Sumanth Udupa\footnotemark[2] $\ ^{1}$, Prajwal Gurunath\footnotemark[2] $\ ^{1}$, Aniruddh Sikdar\footnotemark[2] $\ ^{2}$, Suresh Sundaram$^{1}$        
\\  
$^{1}$Department of Aerospace Engineering, Indian Institute of Science, Bengaluru, India\\
$^{2}$Robert Bosch Centre for Cyber-Physical Systems, Indian Institute of Science, Bengaluru, India\\
\\
{\tt\small \{sumanthudupa, prajwalg, aniruddhss,  vssuresh\} @iisc.ac.in}
}

\begin{document}
\maketitle
\renewcommand{\thefootnote}{\fnsymbol{footnote}}
\footnotetext[2]{Equal contribution of authors.}
\begin{abstract}
Deep neural networks have shown exemplary performance on semantic scene understanding tasks on source domains, but due to the absence of style diversity during training, enhancing performance on unseen target domains using only single source domain data remains a challenging task. Generation of simulated data is a feasible alternative to retrieving large style-diverse real-world datasets as it is a cumbersome and budget-intensive process. However, the large domain-specfic inconsistencies between simulated and real-world data pose a significant generalization challenge in semantic segmentation. In this work, to alleviate this problem, we propose a novel Multi-Resolution Feature Perturbation (MRFP) technique to randomize domain-specific fine-grained features and perturb style of coarse features. Our experimental results on various urban-scene segmentation datasets clearly indicate that, along with the perturbation of style-information, perturbation of fine-feature components is paramount to learn domain invariant robust feature maps for semantic segmentation models. MRFP is a simple and computationally efficient, transferable module with no additional learnable parameters or objective functions, that helps state-of-the-art deep neural networks to learn robust domain invariant features for simulation-to-real semantic segmentation. %Add github link.

\end{abstract}    
\section{Introduction}
\label{sec:intro}

Semantic segmentation is a fundamental computer vision task, with diverse downstream applications, such as autonomous driving \cite{yang2018real}, robot navigation \cite{thomas2021self, panda2023agronav}, medical image analysis \cite{ahn2021spatial}, landcover classification \cite{xia_openearthmap_2023}, and building detection \cite{chen2018encoder, sikdar_deepmao_nodate}. 
Producing synthetic data for training deep neural networks (DNNs) is a cost-effective and straightforward alternative compared to the myriad of challenges associated with collecting real-world data. However, models trained on synthetic data leads to the domain shift problem \cite{ben2010theory,chen2021semantics, luo2019taking, vu2019advent, saito2018maximum}, resulting in a drastic drop in performance.
% The domain shift \cite{ben2010theory,chen2021semantics} that occurs when these models are trained on a given dataset (i.e., source domain) to an unseen dataset (i.e. target domain) results in a drastic drop in performance\cite{pan2018two, luo2019taking, hoffman2018cycada,vu2019advent, saito2018maximum} leading to model uncertainties. 
In safety-critical applications such as autonomous driving, different illuminations, adverse weather conditions and the domain shift from synthetic data \cite{richter2016playing, ros_synthia_2016} cause a significant drop in the performance of DNNs when tested on real-world datasets. There are two primary challenges with training models on synthetic datasets: 1) It is not feasible to synthetically generate all potential unseen domains. 2) Models trained on synthetic data do not generalize to real-world scenarios due to the domain gap that persists when deployed in real-world situations.\\
In this paper, we propose a novel feature perturbation technique referred to as Multi-Resolution Feature Perturbation (MRFP) to address the domain gap between the source and target domains, especially in a Sim-2-Real Single Domain Generalization (SDG) setting. We hypothesise that there exists a latent space consisting of domain-agnostic features that are: 1) independent of style information such as illumination and color characterized as low frequency (LF) components; and 2) fine-grained local texture information characterized as high-frequency (HF) components. The objective of MRFP is to selectively perturb domain-variant LF and HF encoder features, aiming to improve the generalizability of DNNs. MRFP achieves this through two divergent receptive field branches tasked to focus on HF fine-grained features and  LF semantic information. While previous works have focused their efforts on SDG \cite{wang2021learning, fan2021adversarially, wan2022meta, Huang_2023_CVPR, wang2022generalizing, zhou2021domain} and some have tackled the Sim-2-Real \cite{choi2021robustnet, choi2023progressive, 9156428} setting, there is a dearth of literature on the Sim-2-Real problem with an image frequency perspective. To address the domain gap, existing approaches involve enhancing the domain variance of the available source domain data and enriching its representation. This is achieved through methods such as introducing adversarial noise perturbation \cite{chen_center-aware_2023, wang_ffm_2023} or employing style manipulation techniques \cite{zhou2021domain, wang_ffm_2023, fan2022towards}. However, most of these techniques involve intricate training procedures and multiple objective functions.

\begin{figure}[t]
    \centering
    \includegraphics[scale=0.41]{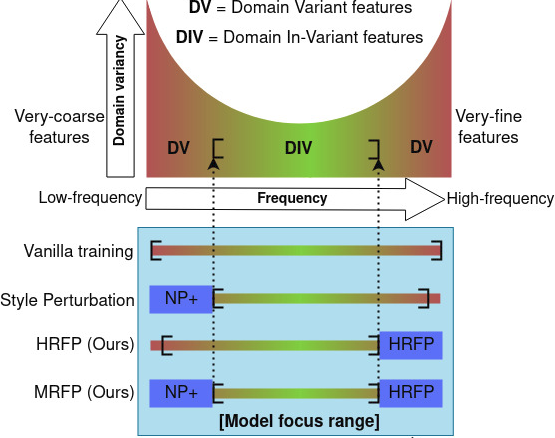}
    \caption{Deep models focus on low-frequency features in the initial stages of vanilla training and shift their focus mainly to domain-variant HF (very-fine) features, covering the entire spectrum. Introducing variability with Style Perturbation (NP+) and High-Resolution Feature Perturbation (HRFP) at both ends of the spectrum, shifts the model's focus to domain in-variant features.}
    \label{fig:Intuition}
\end{figure}

DNNs can focus on a broad spectrum of frequencies, ranging from low to high. Wang \etal  \cite{9156428} suggests that convolution-based DNNs tend to grasp LF attributes in the initial training phases, and gradually transition their attention towards HF components that are notably more domain-specific. This phenomenon is illustrated with \textit{vanilla training} in Fig.\ref{fig:Intuition}. Convolution-based DNNs  progressively cover the entire frequency spectrum as marked by `[ ]' in Fig.\ref{fig:Intuition} denoting the model focus range. This wide range also includes fine-grained domain-specific information. Huang \etal \cite{Huang_2021_CVPR} shows that the lowest and the highest spectral bands in the frequency domain capture domain variant features which hinders generalization performance on unseen domains. 
From our observations of the feature space, it becomes apparent that high-resolution (in a spatial sense) features mostly correlate to fine-grained features (as studied in Section 4). Similarly, we observe that low-resolution (in a spatial sense) features correspond to coarse features. 
Drawing connections from a spatial to a Fourier perspective, we hypothesize that fine-grained information predominantly corresponds to domain-specific HF features. While coarse features correspond to LF features. 

% Hence, overcomplete branches are used to perturb the model at both the shallow and deep layer, to shift the model's focus progressively towards  mid-level domain invariant frequency components. 
% This shift in focus of deep learning models towards domain invariant frequency bands using overcomplete models has not yet been explored for domain generalization.

MRFP not only contributes to style perturbation but also provides control over perturbation of fine-grained features.  It primarily consists of two components, i.e, the High-Resolution Feature Perturbation (HRFP) module, which comprises of a randomly initialized overcomplete (in a spatial sense) autoencoder, and style perturbation with the normalized perturbation technique (NP+) \cite{fan2022towards} within the feature space. Style perturbation techniques at the image level, however, is limited, deterministic and sacrifices source domain performance because of its potential to adversely affect image content \cite{fan2022towards}. Although NP+ (a feature level perturbation technique) can generate diverse styles while preserving high content fidelity, it does not perturb domain-specific fine-grained features, thus missing opportunities to further enhance generalizability. Randomizing these features during the training process restricts the model from drawing inherent source domain specific patterns. 

RandConv \cite{xu2020robust} and ProRandConv \cite{choi2023progressive} are image augmentation techniques that introduce variance in features in the form of contrast and texture diversification in the image space. In contrast to ProRandConv, the proposed HRFP module operates in the feature space by extracting fine-grained characteristics via a decreasing receptive field from a randomly initialized overcomplete autoencoder, serving as perturbations to prevent domain-variant feature overfitting. The inherent benefit of utilizing overcomplete convolutions lies in their decreasing receptive field, where perturbations do not induce significant semantic distortions. This is because the influence of pixels beyond the center of the receptive field is minimal compared to the increased impact observed in undercomplete networks with an increasing receptive field.
As shown in  Fig.\ref{fig:Intuition}, the introduction of \textit{style perturbation} with NP+ and \textit{HRFP} helps restrict the model focus range to domain in-variant features. 

To the best of our knowledge, MRFP is the first technique that employs decreasing receptive fields of overcomplete networks to focus on fine-grained features and perturb them through randomly initialized weights to enhance generalization performance. To summarize, the overall main contributions are as follows:

\begin{itemize}
\item A novel MRFP technique is proposed to introduce perturbations to fine-grained information and infuse varied style information into any baseline segmentation encoder backbone to facilitate the learning of domain-agnostic semantic features.

\item The proposed HRFP technique aims to prevent overfitting on source domain, by using a randomly initialized overcomplete autoencoder on the shallow encoder layers and decoder layer of the baseline segmentation model as a feature-space perturbation. 

\item MRFP is a simple transferable module with no additional learnable parameters or objective functions, which improves the generalizability of deep semantic segmentation models.

\item Extensive experiments over seven urban semantic segmentation datasets show that the proposed model achieves superior performance for single and multi-domain generalization tasks in a Sim-2-Real setting.
\end{itemize}

\section{Related Works}
\hspace{12pt}\textbf{Domain Generalization:} 
These techniques aim to improve the generalization ability of models to unseen target domains, without any access to these domains during training. Various methods like domain alignment \cite{gretton2012kernel}, meta-learning \cite{li2018learning}, adversarial learning \cite{wang_ffm_2023}, and data augmentation \cite{hoffman2018cycada, chen_center-aware_2023} have been proposed to learn domain invariant features. 
IBN-Net \cite{pan2018two} shows significant improvement combining batch and instance norm to learn discriminative features and avoid overfitting on the training data. 
NP+ \cite{fan2022towards} has been used to perturb the feature statistics and synthesize diverse domain styles.
Whitening transform has shown to eliminate style information when applied to each instance \cite{li2017universal}, but may remove domain invariant content at the same time. 
An instance selective whitening loss was proposed by Robustnet \cite{choi2021robustnet} to selectively remove only feature representations that cause domain shifts. 
WildNet \cite{lee2022wildnet}, SANSAW \cite{peng_semantic-aware_2022} and Style Projected Clustering \cite{huang2023style} are recent DG methods that either use external real-world data (ImageNet) for synthesising styles or have complicated training strategies with multiple objective functions. MRFP on the other hand, does not introduce any learnable parameters during the training phase or use any external data.

\textbf{Data Augmentation and Domain Randomization:} 
Data augmentation and domain randomization \cite{tobin2017domain} techniques are used to expand the training data for better generalization. For classification task, frequency based techniques like APR \cite{chen2021amplitude} have been proposed, where the main idea is to augment only the amplitude spectrum of an image  while keeping the phase spectrum constant. Style randomization is used to expand the coverage and diversify the source domain using normalization layers \cite{jackson2019style} and random convolutions \cite{xu2020robust}.  Progressive random convolutions \cite{choi2023progressive}  employ progressively stacked randomly initialized convolutions with an increasing receptive field, as an image-space style perturbation to introduce diversity in style and contrast. 
In contrast, the suggested MRFP technique perturbs both low and high frequencies in the feature space, incorporating both style perturbation and HF perturbation.

\begin{figure*}[ht]
    \centering
    \includegraphics[scale=0.34]{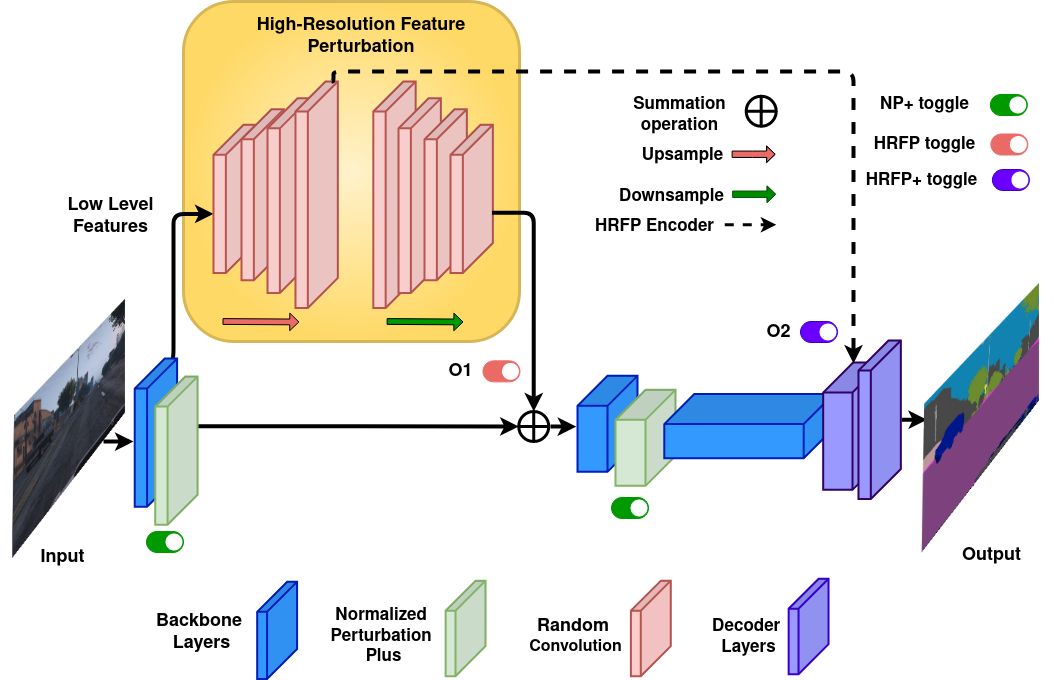}
    \caption{Multi-Resolution Feature Perturbation Technique: Normalized Perturbation (NP+) and High-Resolution Feature Perturbation (HRFP) are randomly incorporated into the training procedure for the baseline segmentation model (DeepLab v3+) , which are represented by the toggles. Dotted line, which is the addition of features to penultimate layer of decoder, is incorporated only in High-Resolution Feature Perturbation Plus (HRFP+) technique. MRFP $\rightarrow$ \{HRFP, NP+\} and MRFP+ $\rightarrow$ \{HRFP, HRFP+, NP+\}}
    \label{fig:MRFP}
\end{figure*}
\textbf{Overcomplete Networks:} Previous studies \cite{jose_kiu-net_2020,sikdar_deepmao_nodate, vincent_extracting_2008} have suggested that overcomplete representations have the ability to pick up on the finer details in the input image while also being robust to noise compared to their undercomplete counterparts in the semantic segmentation task. It is shown that these fine features are paramount for high source domain accuracy \cite{walter2022fragile}. However, the generalization ability of overcomplete models has not been explored. When subjected to diverse domains, learning these fine features can have a detrimental impact on out-of-domain performance. MRFP however, employs these fine features as perturbations to prohibit the model from overfitting on HF domain-specific features.

% \textbf{Overcomplete models.} These representations were explored in signal processing to represent the input signal samples with more number basis functions. It was shown that not only are the overcomplete bases better approximators of the underlying distribution of data but are also more robust to noise\cite{valanarasu2021kiu}. Models like  denoising autoencoder with overcomplete layers performed better as feature detectors\cite{vincent2008extracting}. 
% Overcomplete convolutional networks have been used for SAR despeckling\cite{perera2022sar}. The finer details captured by the overcomplete branch are useful for removing the fine speckles and a Multi-Scale Feature Fusion block is proposed to transfer the low-level features of the overcomplete branch to the under-complete branch. A Fine Context-aware Shadow Detection Network (FCSD-Net) \cite{valanarasu2023fine} was proposed to detect unclear and blurry shadow regions by using upsampling layers to reduce the receptive field as the network propagates deeper.\\
% For the binary pixel classification task of building detection, DeepMAO was proposed. It uses the overcomplete branch to detect finer buildings even in noisy SAR images.\\
% However, to our knowledge, this is the first work
% that uses overcomplete models on diverse datasets to show
% the consistent improvement compared to their undercomplete
% counterparts.\\ It is also the first time that the challenging task of domain generalization has been proposed for overcomplete models.

\section{Methodology}
\subsection{Problem Formulation}

Domain generalization aims to learn a domain-independent model, to train on only a source domain \textit{S} and test on unseen target domains \textit{T} = \{$\textit{T}_{1},\textit{T}_{2}...\textit{T}_{n}$\}. Let the source domain training dataset be denoted as \textit{S} = $\{\textit{x}_{n},\textit{y}_{n}\}_{n=1}^{N_{S}}$ where 
\textit{$x_{n}$} is the $n^{th}$ image, \textit{$y_{n}$} is its corresponding pixel-wise label, and \textit{$N_{s}$} is the total samples in the source domain S. The focus of this work is the semantic segmentation task, with the assumption of a common label space for source and unseen target domains. 
For multi-domain generalization case, where multiple source domains exist, \textit{S} =  \{$\textit{S}_{1},\textit{S}_{2}...\textit{S}_{n}$\} are used during training, and for each training iteration, 
samples are selected randomly from multiple source domains as input. The objective function to train the model using empirical risk minimization \cite{vapnik1999nature} is given as:

\begin{equation} \label{eq:1}
\begin{aligned}
% \argmin_\theta \frac{1}{N_S} \sum_{n=1}^{N_S} l(f_\theta(x_n),y_n)\\
\arg \min_\theta \frac{1}{N_S} \sum_{n=1}^{N_S} l(f_\theta(x_n),y_n)\\
\end{aligned}
\end{equation}
where $f_\theta$$(\cdot)$ is the semantic segmentation network that outputs pixel-wise category predictions, $\theta$ represents the learnable parameters in the network, and \textit{l}$(\cdot)$ is the cross-entropy loss function to measure error. To make the segmentation model $f_\theta$$(\cdot)$ generalizable by learning domain invariant features in both single and multi-domain settings, MRFP technique is proposed.
\subsection{Preliminary: Overcomplete Representations}
The proposed HRFP module is a randomly initialized overcomplete auto-encoder. In this sub-section a brief overview of overcomplete representations is presented.
Let \textit{F1} and \textit{F2} be the feature maps of input image \textit{I}. Assume that the initial Receptive Field (RF) of the convolutional filter is \textit{k $\times$ k} on the image. In undercomplete auto-encoders, due to the max pooling operation, the spatial dimensionality of \textit{F1} is halved causing an increase in the receptive field of \textit{F1, F2} and so on. Eq. \ref{eq:2} is the generalized RF equation for the \textit{i$^{th}$} layer.

\begin{equation} \label{eq:2}
\begin{aligned}
R.F. \hspace{3pt}\textit{w.r.t I} = 2^{2(i-1)} \times k \times k
\end{aligned}
\end{equation}
\noindent However, with overcomplete architectures spatial dimensionality of features F1, F2 and so on increase. For example, an upsampling operation of coefficient 2 that replaces the max pooling operation causes a decrease in the receptive field as generalized in Eq. \ref{eq:3}.
\begin{equation} \label{eq:3}
\begin{aligned}
R.F. \hspace{3pt}\textit{w.r.t I} = (1/2)^{2(i-1)} \times k \times k
\end{aligned}
\end{equation}
\noindent Due to their decreasing receptive fields overcomplete architectures focus on meaningful fine-grained information \cite{jose_kiu-net_2020}.

\subsection{Multi-Resolution Feature Perturbation}

An inherent tactic for addressing domain shift in domain generalization approaches involve producing diverse data and integrating it into the training set. 
In the proposed MRFP technique, in addition to creating diverse styles it aims to perturb the distribution of domain-specific fine-grained features so that the model does not tend to overfit on these fragile features. MRFP technique has two main components namely the high-resolution  feature perturbation module and the NP+ module, which are detailed below. 
%Both components operate on the initial layers of the backbone network of the segmentation model and aim to perturb low-level features but in distinct methods.\\

\textbf{High-Resolution Feature Perturbation (HRFP):}  DNNs overfitting on fine-grained features are shown to be detrimental to performance when tested on unseen domains \cite{9156428}. To address this issue, we employ a randomly initialized overcomplete convolutional auto-encoder that transforms input features to a higher dimension (in a spatial sense). With a decreasing receptive field in HRFP, there exists a high focus on fine-grained features. These fine-grained features are perturbed using random convolutional and batch norm layers as shown in Fig.\hspace{3pt}\ref{fig:MRFP}.\hspace{3pt}These perturbations are subsequently added to the base network to prevent the model from identifying inherent domain-specific patterns.

HRFP is a plug and play module, and can work with any deep segmentation encoder backbone. The encoder and decoder of HRFP consists of 4 convolutional layers each, where every randomly initialized convolutional layer is followed by a randomly initialized batch normalization layer. The reduction in receptive field as denoted in Eq.\hspace{3pt}\ref{eq:3} occurs as a result of increasing the spatial resolution of the feature maps in each layer consecutively by bilinear interpolation with a scaling factor of approximately 1.2 in the HRFP encoder. The HRFP module is upsampled up to a maximum spatial resolution that is twice the size of its own input. Following this, the final four layers of the HRFP module reduce the spatial dimensionality of the overcomplete latent space back to its original input size, facilitating its smooth integration with the base network. Further details are provided in the supplementary material.
% An advantage of the randomly initialized auto-encoder block with decreasing receptive field  is the reduction in the semantic content distortion. 
The random convolution weights are He-initialized \cite{he2015delving} whereas, random batch normalization weights (i.e  $\gamma$ and $\beta$) are sampled from a Gaussian distribution $\cal{N}$(0, $\sigma^2$). The input to the HRFP module originates from the output of the initial stage (stage 0) of the encoder from the backbone layers  of the baseline segmentation network  DeepLabv3+, as shown in Fig. \ref{fig:MRFP}. Since  the shallow layers in CNNs preserve style related information through encoding local structures\cite{10.1007/978-3-319-10590-1_53}, we focus on feature perturbation in these layers, which empiracally yielded the best performance. Hence, the output of HRFP is incorporated as a perturbation, added to the output of the same layer within the base network's encoder backbone as shown in $O_{1}$ branch in Fig.\hspace{3pt}\ref{fig:MRFP}. 

\textbf{HRFP+:} In addition, to further encourage the model to focus towards learning robust domain-invariant features, fine-grained perturbation is added to the decoder of the base network. In HRFP+, the output of the largest upsampled encoder layer of the HRFP block is added to the penultimate layer of the decoder of the baseline segmentation model as shown by branch $O_{2}$ in Fig. \ref{fig:MRFP}. The intuition behind this extra perturbation in the decoder of the base network is that it adversarially helps the segmentation head and makes it more robust against domain-specific HF noise. Additionally in this case, three instance normalization layers \cite{pan2018two} have been adopted in a similar fashion as \cite{choi2021robustnet}.
% \begin{figure}[h]
%     \centering
%     \includegraphics[scale=0.25]{./Figures/NpandHRFP.png}
%     \caption{Presence of frequencies in the Fourier domain of normalized perturbation and HRFP features. The most significant rise in the presence of low-frequency features is noted following NP+ perturbation (left), whereas the most notable increase in the presence of high-frequency features occurs with HRFP perturbation (right). }
%     \label{fig:NPHRFP}
% \end{figure}

\textbf{Style Perturbation:}
From our conjecture that low-resolution features correspond to LF features, we aim to increase the variations in the low-frequencies by perturbing feature channel statistics in the spatial domain. To facilitate an increase in the diversity of style which is known to correspond to LF components in the amplitude spectrum, feature channel statistics in the spatial domain are modified using normalized perturbation \cite{fan2022towards}. NP+ helps the model perceive potentially diverse domains and not overfit to the source domain, and is given by, 
\begin{equation} \label{eq:4}
  \begin{aligned}
    y & = \sigma_{s}^{*} \frac{x-\mu_{c}}{\sigma_{c}}\ + \mu^{*}_{s}, \quad  \sigma_{s}^{*} = \alpha \sigma_{c},
    \quad  \mu_{s}^{*} = \beta \mu_{c}
  \end{aligned}
\end{equation}
where \{$\mu_{c},\sigma_{c}$\} $\in$ $\mathbb{R}^{B \times C}$  are mean and variance of input channels, and  \{$\alpha,\beta$\} $\in$ $\mathbb{R}^{B \times C}$ are drawn from normal distribution. 
For a batch B with feature channel statistics $\mu_{c}$, the statistic variance $\Delta$ $\in$ $\mathbb{R}^{1 \times C}$ is given by,\\
\begin{equation} \label{eq:5}
  \begin{aligned}
    \Delta & =  \frac{1}{B} \sum_{b=1}^{B}(\mu^{b}_{c} - \tilde \mu_{c})^{2}, \quad \tilde \mu_{c} = \frac{1}{B} \sum_{b=1}^{B}(\mu^{b}_{c})
  \end{aligned}
\end{equation} 
where, $\mu^{b}_{c}$ is the feature channel mean of $b^{th}$ sample in the batch. Setting the normalized variance $\delta$ = $\Delta$/max($\Delta$) and using Eq. \ref{eq:4}, the output feature map \textit{y} is given as,
\begin{equation} \label{eq:6}
  \begin{aligned}
    y = \alpha x + \delta(\beta-\alpha)\mu_{c}
  \end{aligned}
\end{equation}
where max represents the maximum operation. Since the features being perturbed using NP+ have a smaller spatial resolution than that of the HRFP block, these perturbations can be viewed as low-resolution feature perturbations. \\

MRFP/MRFP+ consists of both high-resolution feature perturbation (HRFP/HRFP+) and normalized feature perturbation. The perturbations caused due to HRFP/HRFP+ enables the model to learn  domain invariant representations in the learned feature space from distinct domains generated from NP+. 
The proposed method is fundamentally different from previous convolutional randomization techniques \cite{choi2023progressive, xu2020robust, yue2019domain} wherein, the task is to induce various style domains in the image space. In contrast, MRFP/MRFP+ aims to induce domain agnostic model behaviour by not only generating diverse style-information in the feature space but also by not letting the model overfit on HF source domain-specific features.
The proposed module is a simple, computationally efficient, transferable technique, and thus can be attached to any deep backbone network while adding no additional learnable parameters, nor extra objective functions to optimize in the training process of the base network. During inference, the MRFP module is removed, and only the baseline segmentation network is used. 

\begin{figure*}[t]
    \centering
    \includegraphics[scale=0.13]{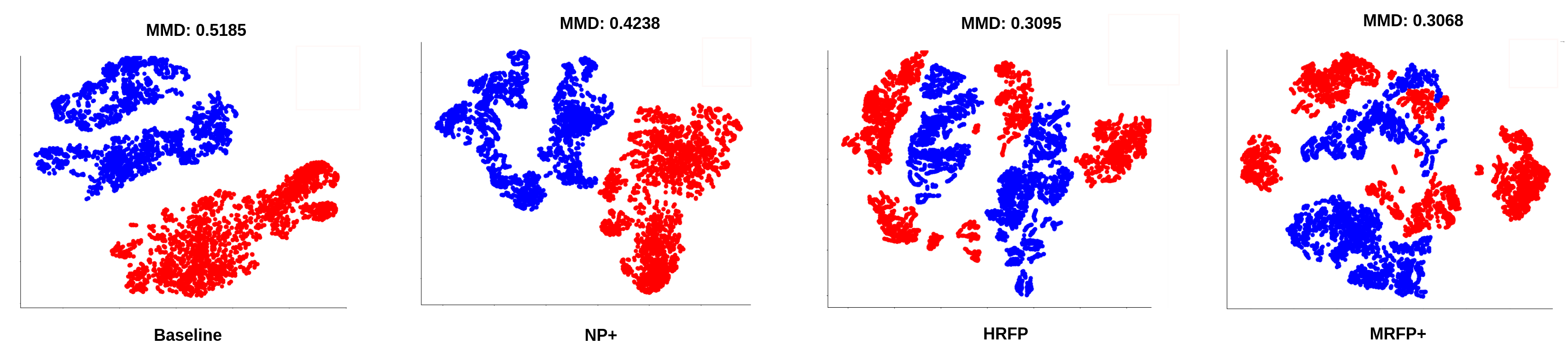}
    \caption{t-SNE visualization for the feature channel statistics of different components of MRFP+, MRFP, NP+ and baseline on GTAV (source domain - red color) and Mapillary (target domain - blue color). The corresponding MMD scores are also reported. }
    \label{fig:TSNE}
\end{figure*}

%\subsection{Training pipeline}
%Mention the training procedure, and explain when the overcomplete branches are removed, and %initialized.

\section{Experiments}
\subsection{Experimental Setup}
The assessment of the proposed MRFP technique involves the use of two synthetic datasets, GTAV \cite{richter2016playing} individually and collectively with Synthia \cite{ros_synthia_2016} to address the challenge of Sim-2-Real domain generalization.  The models that undergo training are subsequently tested on unseen domains that were not part of the training process. These new domains encompass BDD100k (B), Cityscapes (C), Mapillary (M), Foggy Cityscapes (F), and either GTAV (G) or Synthia (S), depending on the specific training configuration.
In scenarios involving single and multi-domain generalization, the setups are as follows: G$\rightarrow$ \{B, C, M, S\}, G$\rightarrow$ \{F, Rainy Cityscapes with intensity level of 25mm, 50mm, 75mm, 100mm\} and (G + S)$\rightarrow$ \{B, C, M\}. In order to ensure fair comparisons, we  re-implement IBN-Net \cite{pan2018two}, RobustNet (ISW) \cite{choi2021robustnet}, and SAN-SAW \cite{peng_semantic-aware_2022}, and evaluate them on F using their open source codes\footnote{Details about re-implemented methods are given in Supplementary material.}.
As explained in Section 3.2, MRFP is a transferable plug-and-play module that can be incorporated into any existing model backbone. Extensive experimentation is carried out using two distinct backbones, namely Resnet-50 and MobileNetv2, showcasing the effectiveness and wide applicability of the proposed module. We use Mean Intersection over Union (mIoU) as our quantitative metric.

\begin{table}[ht]
% \centering
\begin{adjustbox}{max width=0.48\textwidth}
\begin{tabular}{l|l|l|l|l|c}
\toprule
Models(GTAV) &  B     & C     & M        & S      & Avg   \\ \midrule
Baseline     &  31.44 & 34.66 & 32.93  & 25.84 & 31.21         \\ \midrule
IBN-Net \cite{pan2018two}      &  32.30  & 33.85 & 37.75 & 27.90 & 32.95      \\ \midrule
ISW \cite{choi2021robustnet}         & 35.20  & 36.58 & 40.33  & 28.30  & 35.10      \\ \midrule
SAN-SAW \cite{peng_semantic-aware_2022}      & 37.34 & 39.75 & 41.86  & 26.70  & 36.41      \\ \midrule
WildNet* \cite{lee2022wildnet}   &  34.65 & 39.13  & 39.05   & \underline{28.41}   & 35.31            \\ \midrule
WEDGE \cite{kim2023wedge}       & 37.00  & 38.36  & \underline{44.82}  & N/A    & N/A \\ \midrule
DIRL \cite{xu2022dirl}        & \underline{39.15}  & 41.04  & 41.60  & N/A    & N/A \\ \midrule
ProRandConv \cite{choi2023progressive} &  37.03 & \underline{42.36} & 41.63 & 25.52 & 36.63      \\ \midrule
MRFP(Ours)   &  38.80 & 40.25 & 41.96 & 27.37 & \underline{37.09} \\ \midrule
MRFP+(Ours) &    \textbf{39.55} & \textbf{42.40} & \textbf{44.93}  & \textbf{30.22}  & \textbf{39.27}         \\ \bottomrule
\end{tabular}
\end{adjustbox}
\caption{Performance comparison of domain generalization methods in terms of mIoU using ResNet-50 backbone. Models are trained on G, and validated on B, C, M, and S. * indicates that the external dataset (i.e., ImageNet) used in WildNet is replaced   with the source dataset for fair
comparison.}
 \label{table:GTAtable}
\end{table}

%Models (GTAV) & Publication & B     & C     & M     & F     & S          \\ \midrule
%Baseline     &             & 31.44 & 34.66 & 32.93 & 33.68 & 25.83          \\ \midrule
%IBN-Net \textsuperscript{\textdagger}     & ECCV-2018   & 32.3  & 33.85 & 37.75 & 33.18   & \underline{27.9}       \\ \midrule
%ISW   \textsuperscript{\textdagger}         & CVPR-2021   & 35.2  & 36.58 & 40.33 & 36.3  & \textbf{28.3}       \\ \midrule
%SAN-SAW   \textsuperscript{\textdagger}      & CVPR-2022   & 37.34 & 39.75 & 41.86 & 33.8  & 26.7 \\ \midrule
%ProRandConv  & CVPR-2023   & 37.03 & \textbf{42.36} & 41.63 & -     &            \\ \midrule
%WildNet*     & CVPR-2022   & 34.82 & 40.1  & 39.38 & tbu     &            \\ \midrule
%MRFP(Ours) &             & \textbf{38.51} & 39.67 & \textbf{42.67} & \underline{36.9}  & 26.13      \\ \midrule
%MRFP+(Ours) &             & \underline{38.43} & \underline{40.62} & \underline{42.44} & \textbf{38.9}  & 27.76     \\ \bottomrule

% \usepackage[normalem]{ulem}
% \usepackage[normalem]{ulem}

% \usepackage[normalem]{ulem}
% \usepackage[normalem]{ulem}
\begin{table}[ht]
\begin{adjustbox}{max width=0.49\textwidth}
\begin{tabular}{l|l|l|l|l|l|l}
\toprule
Models(GTAV) & F       & 25mm & 50mm & 75mm & 100mm & Avg   \\ \midrule
IBN-Net \cite{pan2018two}     & 33.18       & 20.07       & 14.85       & 11.16       & 8.80          & 17.61 \\ \midrule
ISW \cite{choi2021robustnet}         & 36.30       & 23.7        & 17.93       & 13.53       & 10.39        & 20.37 \\ \midrule
WildNet*\textsuperscript{\textdagger} \cite{lee2022wildnet}     & \underline{38.75}       & 27.04       & 19.17       & 13.94       & 9.45         & 21.67 \\ \midrule
MRFP(Ours)   & 37.90       & \underline{28.29}       & \underline{21.86}       & \underline{17.05}       & \underline{12.79}        & \underline{23.57} \\ \midrule
MRFP+(Ours)  & \textbf{40.67}       & \textbf{30.42}       & \textbf{23.74}       & \textbf{19.25}       & \textbf{15.28}        & \textbf{25.87} \\ \bottomrule
\end{tabular}
\end{adjustbox}
\caption{Performance comparison of domain generalization methods using ResNet-50 backbone, in terms of mIoU. Models are trained on GTAV and tested on adverse weather conditions like fog (foggy Cityscapes) and different levels of rain intensity in mm (rainy cityscapes). \textdagger denotes re-implementation of the method. The best result is highlighted, and the second best result is underlined.}
\label{table:adverse_weather}
\end{table}
% \vspace*{-5mm}
\subsection{Datasets Description}

\textbf{Synthetic Datasets:} 
GTAV \cite{richter2016playing} is a synthetic image dataset generated using the GTA-V game engine, comprising of 24966 images with pixel-wise semantic labels, with a resolution of 1914x1052. Similarly, Synthia \cite{ros_synthia_2016} is also a synthetically generated dataset which includes 9400 images with a resolution of 1280x760. Meanwhile, Synthia has 6580 training images and 2820 validation images. Both datasets have 19 common object categories. 

\noindent \textbf{Real-world Datasets:} 
Five real-world datasets are used, namely Cityscapes (C), Foggy Cityscapes (F), BDD-100k (B), Mapillary (M) and Rainy Cityscapes (R), maintaining a common label space of 19 classes. These datasets are exclusively employed for testing purposes, using their respective validation sets. Cityscapes \cite{cordts2016cityscapes} is a large-scale dataset, with the resolution as 2048x1024, which contains 500 validation samples. In F \cite{sakaridis2018semantic}, synthetic fog is added to the Cityscapes images to simulate reduced visibility conditions, and contains the 1500 images  in validation set based on different foggy settings. BDD-100k \cite{yu2020bdd100k} and Mapillary \cite{neuhold2017mapillary} both contain diverse street view images of resolution 1280x720 and 1920x1080 respectively. The images in validation set of B are 1000 and 2000 for M.
% \vspace{-2mm}
\subsection{Implementation Details}
The baseline segmentation model employed is DeepLabV3+ \cite{chen2018encoder}, implemented with Resnet50 \cite{he2016deep} and MobileNetv2 \cite{sandler2018mobilenetv2} backbones, both utilizing a dilation rate of 16. All backbones used for training are pretrained on ImageNet \cite{deng2009imagenet}. All experiments use SGD \cite{krizhevsky2012imagenet} optimizer with a momentum of 0.9 and  1e-4 as the weight decay. Initial learning rate is set to 0.01 and is decreased according to polynomial rate scheduler with a power of 0.9. All  Resnet50 and MobileNetv2 backbone models are trained for 40k iterations with a batch size of 16 for both single and multi-domain generalization settings. Our augmentation, dataset splits and validation settings are consistent with \cite{choi2021robustnet}. For the randomly initialized HRFP module, the batch-norm layer parameters are sampled from a Gaussian distribution with a standard deviation of 0.5. During the training process, NP+, HRFP and HRFP+ modules are subjected to independent randomization, each with a probability of 0.5.
Inference of the trained models are conducted on the baseline segmentation network DeepLabV3+ only. To ensure equitable evaluations, all the results reported in this subsection are averaged over three separate runs for fair comparisons. Further implementation details are provided in the supplementary material. 

\begin{table}[ht]
\begin{adjustbox}{max width=0.49\textwidth}
\begin{tabular}{l|l|l|l|l|l|l}
\toprule
Models (GTAV)    & B     & C     & S     & M     & F     & Avg   \\ \midrule
Baseline        & 26.76 & 26.95 & 22.72 & 27.34 & 27.26 & 26.20 \\ \midrule
IBN-Net \cite{pan2018two}        & 27.66 & 30.14 & \underline{24.98}  & 27.07 & 30.03     & 27.98      \\ \midrule
ISW \cite{choi2021robustnet}            & \underline{30.05} & \underline{30.86} & 24.43 & \underline{30.67} & \underline{30.70}  & \underline{29.34} \\ \midrule
MRFP (Ours)    & \textbf{33.03} & \textbf{32.92} & \textbf{25.62} & \textbf{30.95} & \textbf{32.97} & \textbf{31.10}  \\ \bottomrule
\end{tabular}
\end{adjustbox}
\caption{Comparison of mIoU (\%). All models are trained on GTAV with DeepLabV3+ with MobileNetv2 as backbone.}
 \label{table:mobilenettable}
\end{table}

% \vspace{-4mm}
\subsection{Experimental Results}
\subsubsection{Quantitative Evaluation}
The domain generalization performance of the proposed MRFP technique is compared with existing methods: IBN-Net \cite{pan2018two}, ISW \cite{choi2021robustnet}, SAN-SAW \cite{peng_semantic-aware_2022}, ProRandConv \cite{choi2023progressive} and WildNet \cite{lee2022wildnet}. 
Table \ref{table:GTAtable} and Table \ref{table:adverse_weather} show the generalization performance  of state-of-the-art (SOTA) models and MRFP with ResNet-50 backbone for single domain generalization, in a Sim-2-Real setting. As shown in Table \ref{table:GTAtable}, both the MRFP and MRFP+ achieve superior performance compared to the SOTA methods on B, C, M and S. It has an improvement of 7.56\% on an average of all target domain datasets compared to baseline DeepLabv3+ model, and an improvement of 2.64\% compared to ProRandConv. Table \ref{table:adverse_weather} shows the generalization performance when trained on G and tested on adverse weather condition datasets like F and R. 
Table \ref{table:mobilenettable} displays the models trained on GTAV, using MobileNetv2 \cite{sandler2018mobilenetv2} as the backbone network. MRFP outperforms all the other methods, showing the plug-and-play nature of the proposed method. 

\begin{table}[ht]
\centering
\begin{adjustbox}{max width=0.60\textwidth}
\begin{tabular}{l|l|l|l|l}
\toprule
Models (G+S)  & B      & C     & M      & Avg   \\ \midrule
Baseline     & 30.11 & 38.63  & 35.90 & 34.88 \\ \midrule
ISW \cite{choi2021robustnet}         & 35.99  & 37.24 & 38.97    & 37.40   \\ \midrule
MRFP (Ours) & \underline{40.35}  & \underline{44.54} & \textbf{45.78} &  \underline{42.55} \\ \midrule
MRFP+ (Ours) & \textbf{41.13}  & \textbf{46.18} & \underline{45.28} &  \textbf{44.24} \\ \bottomrule

\end{tabular}
\end{adjustbox}
\caption{Performance comparison of domain generalization methods in terms of mIoU (\%) using ResNet-50 backbone. Models are trained on (G+S) $\rightarrow$ \{B, C, M \}. }
\label{table:GNS}
\end{table}

Table \ref{table:GNS} shows the Sim-2-Real, multi-domain generalization performance of various domain generalization methods trained on G and S (G+S) datasets combined. The proposed model outperforms the baseline model by 9.36 \% and ISW by 6.84 \%. MRFP+ demonstrates enhanced generalization performance in the Sim-2-Real scenario by enhancing the base network's ability to capture robust and domain-invariant features. In contrast, other approaches primarily focus on normalization, whitening techniques, or introducing random styles. These randomization techniques introduce perturbations with the intention of manipulating input image styles, aiming to cover portions of the target domain. Our method not only covers this aspect, but also induces robustness by perturbing highly domain-specific features in the shallow layers of the encoder. Importantly, this approach doesn't introduce any increase in computational complexity, as only the baseline DeepLabV3+ segmentation model is used during inference. 

\begin{table}[ht]
\begin{adjustbox}{max width=0.48\textwidth}
\begin{tabular}{l|l|l|l|l|l}
\toprule
Models (GTAV)       & B     & C     & M     & S     & Avg    \\ \midrule
Baseline           & 31.44 & 34.66 & 32.93 & 25.83 & 31.71  \\ \midrule
HRFP+     & \underline{39.28} & \underline{41.39}& 42.70 & \underline{29.70} & \underline{38.26}  \\ \midrule
HRFP    & 34.18 & 38.15 & \underline{43.33} & 26.71 & 35.59 \\ \midrule
NP+ & 34.50 & 40.33 & 38.85  & 28.65 &  35.58 \\ \midrule
SCFP & 38.57 & 40.65 & 42.48  & 28.24 & 37.48  \\ \midrule

MRFP(Ours)   &  38.80 & 40.25 & 41.96 & 27.37 & 37.09 \\ \midrule
MRFP+(Ours) &    \textbf{39.55} & \textbf{42.40} & \textbf{44.93}  & \textbf{30.22}  & \textbf{39.27}         \\ \bottomrule
\end{tabular}

\end{adjustbox}
\caption{Ablation analysis of each setting of the MRFP block, mIoU (\%) is reported using ResNet-50 as backbone in the scenario: G $\rightarrow$ \{B, C, M, S\} }
 \label{table:ablationtable}
\end{table}
% \vspace{-6mm}
\subsubsection{Qualitative Evaluation}
\begin{figure}
    \centering
    \includegraphics[scale=0.33]{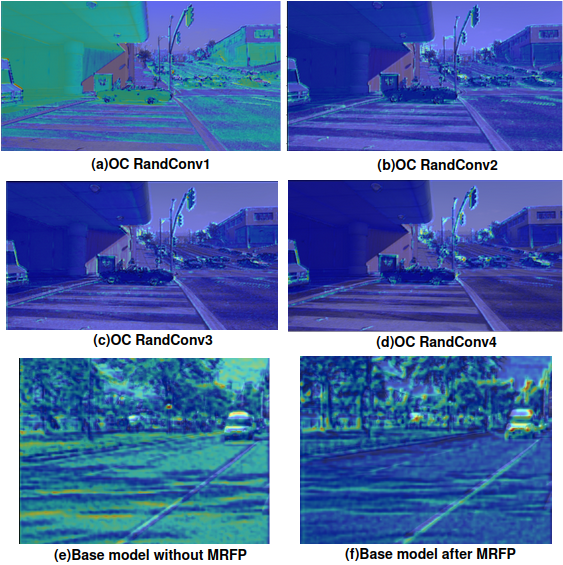}
    \caption{GradCAM outputs of subsequent HRFP layers (a-d), shows that constriction of receptive field, forces the module to focus on fine-grained information. (e) showcases model focus on domain-specific features whereas (f) with MRFP the base model focuses on domain in-variant meaningful features.}
    \label{fig:GRADCAM}
\end{figure}
% \vspace*{-5mm}
\begin{figure}[ht]
    \centering
    \includegraphics[scale=0.28]{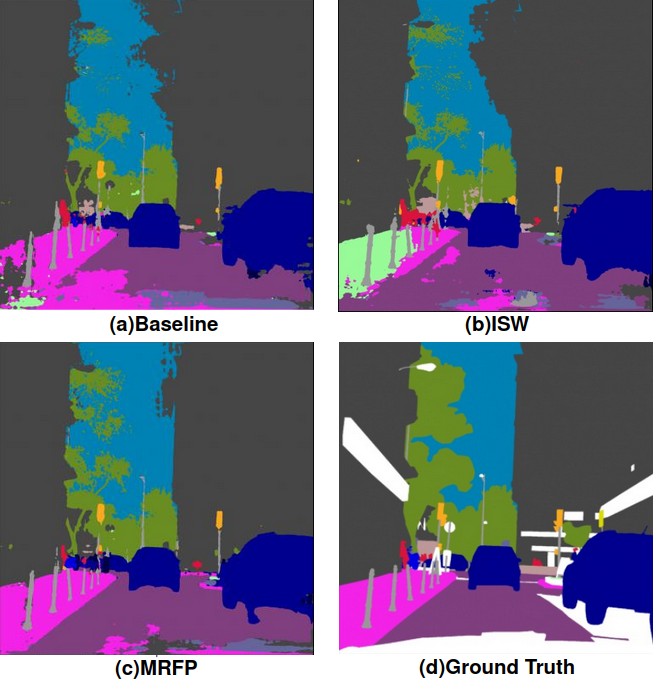}
    \caption{Segmentation outputs of contemporary generalization methods with ground truth.}
    \label{fig:Segmaps}
\end{figure}
To analyze the MRFP module, GradCAM visualizations are shown in Fig. \ref{fig:GRADCAM}. Due to the HRFP block lacking learnable weights, we adopt an ultra-low learning rate in model training for one iteration to create GradCAM visualizations for validating our hypothesis. The focus of the HRFP module moves from coarse to fine features from (a) to (d). It is also seen in (f) that with MRFP in the training procedure, the base model tends to focus more on domain in-variant features such as the vehicle and trees (in this case), as opposed to very-fine textures on the tarmac. 

Fig. \ref{fig:Segmaps} shows the predictions of contemporary generalization models trained on GTAV, and tested on Mapillary. The model extends the source class content to the target domains, e.g, it accurately predicts the pavement, vehicles and roads compared to the baseline and ISW. Further experimental results and segmentation outputs are reported in the supplementary material. 
\subsubsection{Ablation Study}

\begin{figure}[t]
    \centering
    \includegraphics[scale=0.25]{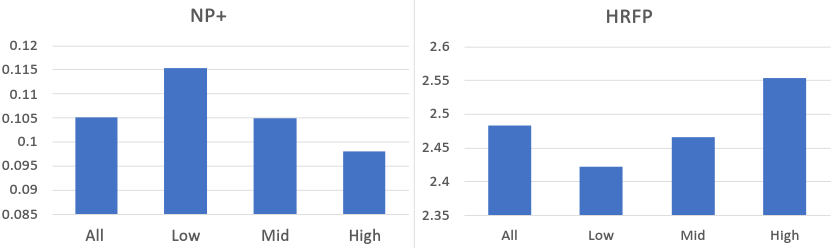}
    \caption{Presence of frequencies in the Fourier domain of NP+ and HRFP features. The most significant rise in the presence of LF features is noted following NP+ perturbation (left), whereas the most notable increase in the presence of high-frequency features occurs with HRFP perturbation (right). }
    \label{fig:NPHRFP}
\end{figure}

 NP+ and HRFP/HRFP+ differ in how they enhance generalization. NP+ targets LF spectrum while HRFP targets the HF spectrum as seen in Fig. \ref{fig:NPHRFP}, which denotes a Fourier analysis of NP+ and HRFP perturbations. A relative increase in the presence of frequencies is studied across 3 bands. The presence of low-frequencies are predominantly increased after NP+. In stark contrast, HRFP predominantly increases the presence of HF components in the feature space. This supports our previously stated conjecture and approach employed to improve out of distribution performance.  
Fig. \ref{fig:TSNE}. depicts the t-SNE plots for the final encoder stage of the ResNet50 backbone along with the corresponding Maximum Mean Discrepancy (MMD) scores. A lower MMD score and a higher degree of overlap between the two distributions indicate better generalizability. This is observed in the components used in the proposed module.

\textbf{Impact of different components in MRFP/MRFP+:} To investigate
the contribution of each component in the overall MRFP technique, we perform an ablation analysis where we validate the need for different components of MRFP as well as different spatial configurations in the proposed HRFP module. Table \ref{table:ablationtable} shows HRFP+ alone outperforms NP+ by 2.68\%, beating SOTA methods by 1.63\%, and the baseline by 6.55\%. Aditionally, NP+ only supplements HRFP/HRFP+, aligning with our hypothesis. All experiments are conducted using the same training settings as described in the implementation details. As seen from the results in Table \ref{table:ablationtable}, we observe that disabling either component has a detrimental effect on out-of-domain performance in comparison to utilizing both components. To further aid the model to focus on domain in-variant features, combining both style perturbation and HRFP+ showcases the highest increase of 7.56\% out-of-domain performance. \\
\hspace{12pt}\textbf{Importance of the overcompleteness in HRFP:} A different spatial configuration in the fine-grained feature perturbation block is conducted with spatially consistent Feature Perturbation (SCFP) where all layers in the fine-grained feature perturbation block are spatially unchanged. The SCFP configuration causes a drop of 0.78\% when compared to HRFP+ module. This could be due to the model's focus not being shifted away from domain-specific fine-grained/HF features. The results show that the combination of HRFP+ and NP+ gives the best average out of domain performance of approximately 7.56\%, showcasing the need for perturbing domain-variant high frequency features along with style perturbation. Additional studies on overcompleteness of the HRFP module are provided in the supplementary material.


%\midrule
%Models(Synthia) & B     & C     & G     & M     & F     & Avg   \\ \midrule
%Baseline        & 17.10 & 23.57 & 26.07 & 19.69 & 19.74 & 21.23 \\ \midrule
%IBN-Net         & tbu   &       &       &       &       &       \\ \midrule
%ISW             & tbu   &       &       &       &       &       \\ \midrule
%MRFP(Ours)    & 18.04 & 28.12 & 27.99 & 22.51 & 23.49 & 24.03 \\ \bottomrule

\section{Conclusion}
This paper introduces a novel Multi-Resolution Feature Perturbation (MRFP) technique, designed to address both single and multi-domain generalization challenges within Sim-2-Real context for semantic segmentation.
The MRFP technique involves perturbing domain-specific fine-grained features using HRFP along with perturbing the feature channel statistics using normalized perturbation.
Our approach consistently achieves superior performance across diverse domain generalization scenarios using seven urban scene segmentation datasets. MRFP has an improvement of 7.56 \% compared to baseline, when trained on GTAV and tested on all target domain datasets. Importantly, this improvement is achieved without an increase in the number of parameters or computational cost during the inference phase.

{
    \small
    \bibliographystyle{ieeenat_fullname}
    \bibliography{main}

\begin{thebibliography}{57}
\providecommand{\natexlab}[1]{#1}
\providecommand{\url}[1]{\texttt{#1}}
\expandafter\ifx\csname urlstyle\endcsname\relax
  \providecommand{\doi}[1]{doi: #1}\else
  \providecommand{\doi}{doi: \begingroup \urlstyle{rm}\Url}\fi

\bibitem[Ahn et~al.(2021)Ahn, Feng, and Kim]{ahn2021spatial}
Euijoon Ahn, Dagan Feng, and Jinman Kim.
\newblock A spatial guided self-supervised clustering network for medical image segmentation.
\newblock In \emph{Medical Image Computing and Computer Assisted Intervention--MICCAI 2021: 24th International Conference, Strasbourg, France, September 27--October 1, 2021, Proceedings, Part I 24}, pages 379--388. Springer, 2021.

\bibitem[Ben-David et~al.(2010)Ben-David, Blitzer, Crammer, Kulesza, Pereira, and Vaughan]{ben2010theory}
Shai Ben-David, John Blitzer, Koby Crammer, Alex Kulesza, Fernando Pereira, and Jennifer~Wortman Vaughan.
\newblock A theory of learning from different domains.
\newblock \emph{Machine learning}, 79:\penalty0 151--175, 2010.

\bibitem[Chen et~al.(2021{\natexlab{a}})Chen, Peng, Ma, Li, Du, and Tian]{chen2021amplitude}
Guangyao Chen, Peixi Peng, Li Ma, Jia Li, Lin Du, and Yonghong Tian.
\newblock Amplitude-phase recombination: Rethinking robustness of convolutional neural networks in frequency domain.
\newblock In \emph{Proceedings of the IEEE/CVF International Conference on Computer Vision}, pages 458--467, 2021{\natexlab{a}}.

\bibitem[Chen et~al.(2018)Chen, Zhu, Papandreou, Schroff, and Adam]{chen2018encoder}
Liang-Chieh Chen, Yukun Zhu, George Papandreou, Florian Schroff, and Hartwig Adam.
\newblock Encoder-decoder with atrous separable convolution for semantic image segmentation.
\newblock In \emph{Proceedings of the European conference on computer vision (ECCV)}, pages 801--818, 2018.

\bibitem[Chen et~al.(2023)Chen, Baktashmotlagh, Wang, and Salzmann]{chen_center-aware_2023}
Tianle Chen, Mahsa Baktashmotlagh, Zijian Wang, and Mathieu Salzmann.
\newblock Center-aware {Adversarial} {Augmentation} for {Single} {Domain} {Generalization}.
\newblock In \emph{2023 {IEEE}/{CVF} {Winter} {Conference} on {Applications} of {Computer} {Vision} ({WACV})}, pages 4146--4154, Waikoloa, HI, USA, 2023. IEEE.

\bibitem[Chen et~al.(2021{\natexlab{b}})Chen, Luo, Qiu, Wang, Huang, Li, and Zhang]{chen2021semantics}
Zhi Chen, Yadan Luo, Ruihong Qiu, Sen Wang, Zi Huang, Jingjing Li, and Zheng Zhang.
\newblock Semantics disentangling for generalized zero-shot learning.
\newblock In \emph{Proceedings of the IEEE/CVF international conference on computer vision}, pages 8712--8720, 2021{\natexlab{b}}.

\bibitem[Choi et~al.(2021)Choi, Jung, Yun, Kim, Kim, and Choo]{choi2021robustnet}
Sungha Choi, Sanghun Jung, Huiwon Yun, Joanne~T Kim, Seungryong Kim, and Jaegul Choo.
\newblock Robustnet: Improving domain generalization in urban-scene segmentation via instance selective whitening.
\newblock In \emph{Proceedings of the IEEE/CVF Conference on Computer Vision and Pattern Recognition}, pages 11580--11590, 2021.

\bibitem[Choi et~al.(2023)Choi, Das, Choi, Yang, Park, and Yun]{choi2023progressive}
Seokeon Choi, Debasmit Das, Sungha Choi, Seunghan Yang, Hyunsin Park, and Sungrack Yun.
\newblock Progressive random convolutions for single domain generalization.
\newblock In \emph{Proceedings of the IEEE/CVF Conference on Computer Vision and Pattern Recognition}, pages 10312--10322, 2023.

\bibitem[Cordts et~al.(2016)Cordts, Omran, Ramos, Rehfeld, Enzweiler, Benenson, Franke, Roth, and Schiele]{cordts2016cityscapes}
Marius Cordts, Mohamed Omran, Sebastian Ramos, Timo Rehfeld, Markus Enzweiler, Rodrigo Benenson, Uwe Franke, Stefan Roth, and Bernt Schiele.
\newblock The cityscapes dataset for semantic urban scene understanding.
\newblock In \emph{Proceedings of the IEEE conference on computer vision and pattern recognition}, pages 3213--3223, 2016.

\bibitem[Deng et~al.(2009)Deng, Dong, Socher, Li, Li, and Fei-Fei]{deng2009imagenet}
Jia Deng, Wei Dong, Richard Socher, Li-Jia Li, Kai Li, and Li Fei-Fei.
\newblock Imagenet: A large-scale hierarchical image database.
\newblock In \emph{2009 IEEE conference on computer vision and pattern recognition}, pages 248--255. Ieee, 2009.

\bibitem[Fan et~al.(2022)Fan, Segu, Tai, Yu, Tang, Schiele, and Dai]{fan2022towards}
Qi Fan, Mattia Segu, Yu-Wing Tai, Fisher Yu, Chi-Keung Tang, Bernt Schiele, and Dengxin Dai.
\newblock Towards robust object detection invariant to real-world domain shifts.
\newblock In \emph{The Eleventh International Conference on Learning Representations}, 2022.

\bibitem[Fan et~al.(2021)Fan, Wang, Ke, Yang, Gong, and Zhou]{fan2021adversarially}
Xinjie Fan, Qifei Wang, Junjie Ke, Feng Yang, Boqing Gong, and Mingyuan Zhou.
\newblock Adversarially adaptive normalization for single domain generalization.
\newblock In \emph{Proceedings of the IEEE/CVF conference on Computer Vision and Pattern Recognition}, pages 8208--8217, 2021.

\bibitem[Gretton et~al.(2012)Gretton, Borgwardt, Rasch, Sch{\"o}lkopf, and Smola]{gretton2012kernel}
Arthur Gretton, Karsten~M Borgwardt, Malte~J Rasch, Bernhard Sch{\"o}lkopf, and Alexander Smola.
\newblock A kernel two-sample test.
\newblock \emph{The Journal of Machine Learning Research}, 13\penalty0 (1):\penalty0 723--773, 2012.

\bibitem[He et~al.(2015)He, Zhang, Ren, and Sun]{he2015delving}
Kaiming He, Xiangyu Zhang, Shaoqing Ren, and Jian Sun.
\newblock Delving deep into rectifiers: Surpassing human-level performance on imagenet classification.
\newblock In \emph{Proceedings of the IEEE international conference on computer vision}, pages 1026--1034, 2015.

\bibitem[He et~al.(2016)He, Zhang, Ren, and Sun]{he2016deep}
Kaiming He, Xiangyu Zhang, Shaoqing Ren, and Jian Sun.
\newblock Deep residual learning for image recognition.
\newblock In \emph{Proceedings of the IEEE conference on computer vision and pattern recognition}, pages 770--778, 2016.

\bibitem[Hoffman et~al.(2018)Hoffman, Tzeng, Park, Zhu, Isola, Saenko, Efros, and Darrell]{hoffman2018cycada}
Judy Hoffman, Eric Tzeng, Taesung Park, Jun-Yan Zhu, Phillip Isola, Kate Saenko, Alexei Efros, and Trevor Darrell.
\newblock Cycada: Cycle-consistent adversarial domain adaptation.
\newblock In \emph{International conference on machine learning}, pages 1989--1998. Pmlr, 2018.

\bibitem[Huang et~al.(2021)Huang, Guan, Xiao, and Lu]{Huang_2021_CVPR}
Jiaxing Huang, Dayan Guan, Aoran Xiao, and Shijian Lu.
\newblock Fsdr: Frequency space domain randomization for domain generalization.
\newblock In \emph{Proceedings of the IEEE/CVF Conference on Computer Vision and Pattern Recognition (CVPR)}, pages 6891--6902, 2021.

\bibitem[Huang et~al.(2023{\natexlab{a}})Huang, Chen, Li, Li, Li, Song, Yan, and Xiong]{Huang_2023_CVPR}
Wei Huang, Chang Chen, Yong Li, Jiacheng Li, Cheng Li, Fenglong Song, Youliang Yan, and Zhiwei Xiong.
\newblock Style projected clustering for domain generalized semantic segmentation.
\newblock In \emph{Proceedings of the IEEE/CVF Conference on Computer Vision and Pattern Recognition (CVPR)}, pages 3061--3071, 2023{\natexlab{a}}.

\bibitem[Huang et~al.(2023{\natexlab{b}})Huang, Chen, Li, Li, Li, Song, Yan, and Xiong]{huang2023style}
Wei Huang, Chang Chen, Yong Li, Jiacheng Li, Cheng Li, Fenglong Song, Youliang Yan, and Zhiwei Xiong.
\newblock Style projected clustering for domain generalized semantic segmentation.
\newblock In \emph{Proceedings of the IEEE/CVF Conference on Computer Vision and Pattern Recognition}, pages 3061--3071, 2023{\natexlab{b}}.

\bibitem[Jackson et~al.(2019)Jackson, Abarghouei, Bonner, Breckon, and Obara]{jackson2019style}
Philip~TG Jackson, Amir~Atapour Abarghouei, Stephen Bonner, Toby~P Breckon, and Boguslaw Obara.
\newblock Style augmentation: data augmentation via style randomization.
\newblock In \emph{CVPR workshops}, pages 10--11, 2019.

\bibitem[Jose et~al.(2020)Jose, Sindagi, Hacihaliloglu, and Patel]{jose_kiu-net_2020}
Jeya~Maria Jose, Vishwanath Sindagi, Ilker Hacihaliloglu, and Vishal~M. Patel.
\newblock {KiU}-{Net}: {Towards} {Accurate} {Segmentation} of {Biomedical} {Images} using {Over}-complete {Representations}, 2020.
\newblock arXiv:2006.04878 [cs, eess].

\bibitem[Kim et~al.(2023)Kim, Son, Pahk, Lan, Zeng, and Kwak]{kim2023wedge}
Namyup Kim, Taeyoung Son, Jaehyun Pahk, Cuiling Lan, Wenjun Zeng, and Suha Kwak.
\newblock Wedge: web-image assisted domain generalization for semantic segmentation.
\newblock In \emph{2023 IEEE International Conference on Robotics and Automation (ICRA)}, pages 9281--9288. IEEE, 2023.

\bibitem[Krizhevsky et~al.(2012)Krizhevsky, Sutskever, and Hinton]{krizhevsky2012imagenet}
Alex Krizhevsky, Ilya Sutskever, and Geoffrey~E Hinton.
\newblock Imagenet classification with deep convolutional neural networks.
\newblock \emph{Advances in neural information processing systems}, 25, 2012.

\bibitem[Lee et~al.(2022)Lee, Seong, Lee, and Kim]{lee2022wildnet}
Suhyeon Lee, Hongje Seong, Seongwon Lee, and Euntai Kim.
\newblock Wildnet: Learning domain generalized semantic segmentation from the wild.
\newblock In \emph{Proceedings of the IEEE/CVF Conference on Computer Vision and Pattern Recognition}, pages 9936--9946, 2022.

\bibitem[Li et~al.(2018)Li, Yang, Song, and Hospedales]{li2018learning}
Da Li, Yongxin Yang, Yi-Zhe Song, and Timothy Hospedales.
\newblock Learning to generalize: Meta-learning for domain generalization.
\newblock In \emph{Proceedings of the AAAI conference on artificial intelligence}, 2018.

\bibitem[Li et~al.(2017)Li, Fang, Yang, Wang, Lu, and Yang]{li2017universal}
Yijun Li, Chen Fang, Jimei Yang, Zhaowen Wang, Xin Lu, and Ming-Hsuan Yang.
\newblock Universal style transfer via feature transforms.
\newblock \emph{Advances in neural information processing systems}, 30, 2017.

\bibitem[Luo et~al.(2019)Luo, Zheng, Guan, Yu, and Yang]{luo2019taking}
Yawei Luo, Liang Zheng, Tao Guan, Junqing Yu, and Yi Yang.
\newblock Taking a closer look at domain shift: Category-level adversaries for semantics consistent domain adaptation.
\newblock In \emph{Proceedings of the IEEE/CVF conference on computer vision and pattern recognition}, pages 2507--2516, 2019.

\bibitem[Neuhold et~al.(2017)Neuhold, Ollmann, Rota~Bulo, and Kontschieder]{neuhold2017mapillary}
Gerhard Neuhold, Tobias Ollmann, Samuel Rota~Bulo, and Peter Kontschieder.
\newblock The mapillary vistas dataset for semantic understanding of street scenes.
\newblock In \emph{Proceedings of the IEEE international conference on computer vision}, pages 4990--4999, 2017.

\bibitem[Pan et~al.(2018)Pan, Luo, Shi, and Tang]{pan2018two}
Xingang Pan, Ping Luo, Jianping Shi, and Xiaoou Tang.
\newblock Two at once: Enhancing learning and generalization capacities via ibn-net.
\newblock In \emph{Proceedings of the European Conference on Computer Vision (ECCV)}, pages 464--479, 2018.

\bibitem[Panda et~al.(2023)Panda, Lee, and Jawed]{panda2023agronav}
Shivam~K Panda, Yongkyu Lee, and M~Khalid Jawed.
\newblock Agronav: Autonomous navigation framework for agricultural robots and vehicles using semantic segmentation and semantic line detection.
\newblock In \emph{Proceedings of the IEEE/CVF Conference on Computer Vision and Pattern Recognition}, pages 6271--6280, 2023.

\bibitem[Peng et~al.(2022)Peng, Lei, Hayat, Guo, and Li]{peng_semantic-aware_2022}
Duo Peng, Yinjie Lei, Munawar Hayat, Yulan Guo, and Wen Li.
\newblock Semantic-{Aware} {Domain} {Generalized} {Segmentation}.
\newblock In \emph{2022 {IEEE}/{CVF} {Conference} on {Computer} {Vision} and {Pattern} {Recognition} ({CVPR})}, pages 2584--2595, New Orleans, LA, USA, 2022. IEEE.

\bibitem[Richter et~al.(2016)Richter, Vineet, Roth, and Koltun]{richter2016playing}
Stephan~R Richter, Vibhav Vineet, Stefan Roth, and Vladlen Koltun.
\newblock Playing for data: Ground truth from computer games.
\newblock In \emph{Computer Vision--ECCV 2016: 14th European Conference, Amsterdam, The Netherlands, October 11-14, 2016, Proceedings, Part II 14}, pages 102--118. Springer, 2016.

\bibitem[Ronneberger et~al.(2015)Ronneberger, Fischer, and Brox]{ronneberger2015u}
Olaf Ronneberger, Philipp Fischer, and Thomas Brox.
\newblock U-net: Convolutional networks for biomedical image segmentation.
\newblock In \emph{Medical Image Computing and Computer-Assisted Intervention--MICCAI 2015: 18th International Conference, Munich, Germany, October 5-9, 2015, Proceedings, Part III 18}, pages 234--241. Springer, 2015.

\bibitem[Ros et~al.(2016)Ros, Sellart, Materzynska, Vazquez, and Lopez]{ros_synthia_2016}
German Ros, Laura Sellart, Joanna Materzynska, David Vazquez, and Antonio~M. Lopez.
\newblock The {SYNTHIA} {Dataset}: {A} {Large} {Collection} of {Synthetic} {Images} for {Semantic} {Segmentation} of {Urban} {Scenes}.
\newblock In \emph{2016 {IEEE} {Conference} on {Computer} {Vision} and {Pattern} {Recognition} ({CVPR})}, pages 3234--3243, Las Vegas, NV, USA, 2016. IEEE.

\bibitem[Saito et~al.(2018)Saito, Watanabe, Ushiku, and Harada]{saito2018maximum}
Kuniaki Saito, Kohei Watanabe, Yoshitaka Ushiku, and Tatsuya Harada.
\newblock Maximum classifier discrepancy for unsupervised domain adaptation.
\newblock In \emph{Proceedings of the IEEE conference on computer vision and pattern recognition}, pages 3723--3732, 2018.

\bibitem[Sakaridis et~al.(2018)Sakaridis, Dai, and Van~Gool]{sakaridis2018semantic}
Christos Sakaridis, Dengxin Dai, and Luc Van~Gool.
\newblock Semantic foggy scene understanding with synthetic data.
\newblock \emph{International Journal of Computer Vision}, 126:\penalty0 973--992, 2018.

\bibitem[Sandler et~al.(2018)Sandler, Howard, Zhu, Zhmoginov, and Chen]{sandler2018mobilenetv2}
Mark Sandler, Andrew Howard, Menglong Zhu, Andrey Zhmoginov, and Liang-Chieh Chen.
\newblock Mobilenetv2: Inverted residuals and linear bottlenecks.
\newblock In \emph{Proceedings of the IEEE conference on computer vision and pattern recognition}, pages 4510--4520, 2018.

\bibitem[Sikdar et~al.(2023)Sikdar, Udupa, Gurunath, and Sundaram]{sikdar_deepmao_nodate}
Aniruddh Sikdar, Sumanth Udupa, Prajwal Gurunath, and Suresh Sundaram.
\newblock Deepmao: Deep multi-scale aware overcomplete network for building segmentation in satellite imagery.
\newblock In \emph{2023 IEEE/CVF Conference on Computer Vision and Pattern Recognition Workshops (CVPRW)}, pages 487--496, 2023.

\bibitem[Thomas et~al.(2021)Thomas, Agro, Gridseth, Zhang, and Barfoot]{thomas2021self}
Hugues Thomas, Ben Agro, Mona Gridseth, Jian Zhang, and Timothy~D Barfoot.
\newblock Self-supervised learning of lidar segmentation for autonomous indoor navigation.
\newblock In \emph{2021 IEEE International Conference on Robotics and Automation (ICRA)}, pages 14047--14053. IEEE, 2021.

\bibitem[Tobin et~al.(2017)Tobin, Fong, Ray, Schneider, Zaremba, and Abbeel]{tobin2017domain}
Josh Tobin, Rachel Fong, Alex Ray, Jonas Schneider, Wojciech Zaremba, and Pieter Abbeel.
\newblock Domain randomization for transferring deep neural networks from simulation to the real world.
\newblock In \emph{2017 IEEE/RSJ international conference on intelligent robots and systems (IROS)}, pages 23--30. IEEE, 2017.

\bibitem[Vapnik(1999)]{vapnik1999nature}
Vladimir Vapnik.
\newblock \emph{The nature of statistical learning theory}.
\newblock Springer science \& business media, 1999.

\bibitem[Vincent et~al.(2008)Vincent, Larochelle, Bengio, and Manzagol]{vincent_extracting_2008}
Pascal Vincent, Hugo Larochelle, Yoshua Bengio, and Pierre-Antoine Manzagol.
\newblock Extracting and composing robust features with denoising autoencoders.
\newblock In \emph{Proceedings of the 25th international conference on {Machine} learning - {ICML} '08}, pages 1096--1103, Helsinki, Finland, 2008. ACM Press.

\bibitem[Vu et~al.(2019)Vu, Jain, Bucher, Cord, and P{\'e}rez]{vu2019advent}
Tuan-Hung Vu, Himalaya Jain, Maxime Bucher, Matthieu Cord, and Patrick P{\'e}rez.
\newblock Advent: Adversarial entropy minimization for domain adaptation in semantic segmentation.
\newblock In \emph{Proceedings of the IEEE/CVF conference on computer vision and pattern recognition}, pages 2517--2526, 2019.

\bibitem[Walter et~al.(2022)Walter, Stutz, and Schiele]{walter2022fragile}
Nils~Philipp Walter, David Stutz, and Bernt Schiele.
\newblock On fragile features and batch normalization in adversarial training.
\newblock \emph{arXiv preprint arXiv:2204.12393}, 2022.

\bibitem[Wan et~al.(2022)Wan, Shen, Zhang, Yin, Tian, Gao, Huang, and Hua]{wan2022meta}
Chaoqun Wan, Xu Shen, Yonggang Zhang, Zhiheng Yin, Xinmei Tian, Feng Gao, Jianqiang Huang, and Xian-Sheng Hua.
\newblock Meta convolutional neural networks for single domain generalization.
\newblock In \emph{Proceedings of the IEEE/CVF Conference on Computer Vision and Pattern Recognition}, pages 4682--4691, 2022.

\bibitem[Wang et~al.(2020)Wang, Wu, Huang, and Xing]{9156428}
Haohan Wang, Xindi Wu, Zeyi Huang, and Eric~P. Xing.
\newblock High-frequency component helps explain the generalization of convolutional neural networks.
\newblock In \emph{2020 IEEE/CVF Conference on Computer Vision and Pattern Recognition (CVPR)}, pages 8681--8691, 2020.

\bibitem[Wang et~al.(2022)Wang, Lan, Liu, Ouyang, Qin, Lu, Chen, Zeng, and Yu]{wang2022generalizing}
Jindong Wang, Cuiling Lan, Chang Liu, Yidong Ouyang, Tao Qin, Wang Lu, Yiqiang Chen, Wenjun Zeng, and Philip Yu.
\newblock Generalizing to unseen domains: A survey on domain generalization.
\newblock \emph{IEEE Transactions on Knowledge and Data Engineering}, 2022.

\bibitem[Wang et~al.(2021)Wang, Luo, Qiu, Huang, and Baktashmotlagh]{wang2021learning}
Zijian Wang, Yadan Luo, Ruihong Qiu, Zi Huang, and Mahsa Baktashmotlagh.
\newblock Learning to diversify for single domain generalization.
\newblock In \emph{Proceedings of the IEEE/CVF International Conference on Computer Vision}, pages 834--843, 2021.

\bibitem[Wang et~al.(2023)Wang, Luo, Huang, and Baktashmotlagh]{wang_ffm_2023}
Zijian Wang, Yadan Luo, Zi Huang, and Mahsa Baktashmotlagh.
\newblock {FFM}: {Injecting} {Out}-of-{Domain} {Knowledge} via {Factorized} {Frequency} {Modification}.
\newblock In \emph{2023 {IEEE}/{CVF} {Winter} {Conference} on {Applications} of {Computer} {Vision} ({WACV})}, pages 4124--4133, Waikoloa, HI, USA, 2023. IEEE.

\bibitem[Xia et~al.(2023)Xia, Yokoya, Adriano, and Broni-Bediako]{xia_openearthmap_2023}
Junshi Xia, Naoto Yokoya, Bruno Adriano, and Clifford Broni-Bediako.
\newblock {OpenEarthMap}: {A} {Benchmark} {Dataset} for {Global} {High}-{Resolution} {Land} {Cover} {Mapping}.
\newblock In \emph{2023 {IEEE}/{CVF} {Winter} {Conference} on {Applications} of {Computer} {Vision} ({WACV})}, pages 6243--6253, Waikoloa, HI, USA, 2023. IEEE.

\bibitem[Xu et~al.(2022)Xu, Yao, Jiang, Jiang, Chu, Han, Zhang, Wang, and Tai]{xu2022dirl}
Qi Xu, Liang Yao, Zhengkai Jiang, Guannan Jiang, Wenqing Chu, Wenhui Han, Wei Zhang, Chengjie Wang, and Ying Tai.
\newblock Dirl: Domain-invariant representation learning for generalizable semantic segmentation.
\newblock In \emph{Proceedings of the AAAI Conference on Artificial Intelligence}, pages 2884--2892, 2022.

\bibitem[Xu et~al.(2020)Xu, Liu, Yang, Raffel, and Niethammer]{xu2020robust}
Zhenlin Xu, Deyi Liu, Junlin Yang, Colin Raffel, and Marc Niethammer.
\newblock Robust and generalizable visual representation learning via random convolutions.
\newblock \emph{arXiv preprint arXiv:2007.13003}, 2020.

\bibitem[Yang et~al.(2018)Yang, Liang, Wang, and Xing]{yang2018real}
Luona Yang, Xiaodan Liang, Tairui Wang, and Eric Xing.
\newblock Real-to-virtual domain unification for end-to-end autonomous driving.
\newblock In \emph{Proceedings of the European conference on computer vision (ECCV)}, pages 530--545, 2018.

\bibitem[Yu et~al.(2020)Yu, Chen, Wang, Xian, Chen, Liu, Madhavan, and Darrell]{yu2020bdd100k}
Fisher Yu, Haofeng Chen, Xin Wang, Wenqi Xian, Yingying Chen, Fangchen Liu, Vashisht Madhavan, and Trevor Darrell.
\newblock Bdd100k: A diverse driving dataset for heterogeneous multitask learning.
\newblock In \emph{Proceedings of the IEEE/CVF conference on computer vision and pattern recognition}, pages 2636--2645, 2020.

\bibitem[Yue et~al.(2019)Yue, Zhang, Zhao, Sangiovanni-Vincentelli, Keutzer, and Gong]{yue2019domain}
Xiangyu Yue, Yang Zhang, Sicheng Zhao, Alberto Sangiovanni-Vincentelli, Kurt Keutzer, and Boqing Gong.
\newblock Domain randomization and pyramid consistency: Simulation-to-real generalization without accessing target domain data.
\newblock In \emph{Proceedings of the IEEE/CVF International Conference on Computer Vision}, pages 2100--2110, 2019.

\bibitem[Zeiler and Fergus(2014)]{10.1007/978-3-319-10590-1_53}
Matthew~D. Zeiler and Rob Fergus.
\newblock Visualizing and understanding convolutional networks.
\newblock In \emph{Computer Vision -- ECCV 2014}, pages 818--833, Cham, 2014. Springer International Publishing.

\bibitem[Zhou et~al.(2021)Zhou, Yang, Qiao, and Xiang]{zhou2021domain}
Kaiyang Zhou, Yongxin Yang, Yu Qiao, and Tao Xiang.
\newblock Domain generalization with mixstyle.
\newblock \emph{arXiv preprint arXiv:2104.02008}, 2021.

\end{thebibliography}
}

\clearpage
\setcounter{page}{1}
\maketitlesupplementary

\section{Supplementary Material}
%%%%%%%%% BODY TEXT - ENTER YOUR RESPONSE BELOW
This supplementary section provides additional details which could not be added in the main paper due to space constrains, including: (1) additional implementation details, (2) additional quantitative experiments, (3) qualitative results, (4) further ablation study, (5) details about computational complexity, and (6) comparison to state of the art domain generalization techniques.

\subsection{Additional Implementation Details}

Each of our models are trained for 40K iterations for both single-domain and multi-domain (G+S) generalization settings. The dataset partitioning (training, validation and test splits) strictly follow \cite{choi2021robustnet}. The re-implementations of other methods also follow the same dataset partitioning. To ensure fair-comparison, \cite{peng_semantic-aware_2022} is also validated on the the Synthia dataset partitioning  of \cite{choi2021robustnet}\footnote{RobustNet data partitioning: \url{http://github.com/shachoi/RobustNet/blob/main/split_data/synthia_split_val.txt}}. The model saved at the last iteration is used for all our analyses for out-of-domain and in-domain performance. The probabilities \textit{p} of all our perturbations (HRFP, HRFP+ and NP+) are empirically set to 0.5 where each probability is independent of another. In short, the perturbations occur on independent probabilities. 
For the style perturbation technique, the noise variance is drawn from a Gaussian distribution of mean 1 and standard deviation on 0.75. The batch size was kept as 16 for all the experiments that used GTAV dataset as the source-domain data. In the case of Synthia dataset being the source domain data, the batch size was kept as 8. All experiments with multi-domain generalization setting was conducted with the batch size being 8.
All models are trained and tested on Nvidia RTX A6000 GPU.
%-------------------------------------------------------------------------
\subsection{Quantitative Results}
Table \ref{table:SynthiaTable} presents the out-of-domain (OOD) performance of MRFP when trained on the Synthia dataset. MRFP improves on the baseline and outperforms other state-of-the-art DG methods by approximately 2\% and 1\% respectively showcasing the wide applicability of the technique for sim-2-real DG. Table \ref{table:SourceG} and Table \ref{table:SourceS} show the source domain performance on  GTAV(G) and Synthia (S) simulated datasets respectively. Although a performance drop of 1.75\% on average is observed in source domain performance compared to the baseline, it is not a major concern, as the overall generalization ability on all  real-world datasets is improved significantly.
Table \ref{table:Unet} displays the wide applicability of the proposed MRFP technique as a plug-and-play module as it improves the OOD performance of a different segmentation network(in this case U-Net \cite{ronneberger2015u} with ResNet-50 as the encoder backbone), without any additional tuning of the hyperparameters by 2.29\%.
\begin{table}[ht]
\centering
\begin{tabular}{l|l|l|l|l|l}
\toprule
Models (S) & B     & C     & M      & G     & Avg   \\ \midrule
Baseline        & 20.83 &  \underline{27.95}  & 24.06  & \underline{29.81} & 25.66 \\ \midrule
IBN-Net \cite{pan2018two}   \textsuperscript{\textdagger}       & 21.12   & 27.09      &   22.94       &  27.38     &    24.63   \\ \midrule
ISW \cite{choi2021robustnet}    \textsuperscript{\textdagger}       & \underline{22.66}   & \textbf{29.92}      & \underline{25.44}         &  28.30     &  \underline{26.58}     \\ \midrule
WildNet*    \textsuperscript{\textdagger} \cite{lee2022wildnet}     & 20.76   &  27.54     &  24.65      &  29.73     &  25.67    \\ \midrule
MRFP (Ours)    & \textbf{25.68} & 27.89 & \textbf{25.98} & \textbf{30.50}  & \textbf{27.51} \\ \bottomrule
\end{tabular}
\caption{Performance comparison of domain generalization methods using ResNet-50 backbone, in terms of mIoU. Models are trained on S $\rightarrow$ \{B, C, M, G \}. \textdagger denotes re-implementation of the method. * indicates that the external dataset(i.e., ImageNet) used in WildNet is replaced with the source dataset for fair comparison. The best result is highlighted, and the second best result is underlined.}
\label{table:SynthiaTable}
\end{table}

\begin{table}[ht]
\centering
\begin{tabular}{|c|c|}
\hline
Models (G)                         & G                          \\ \hline
Baseline                           & 76.57                      \\ \hline
ISW \cite{choi2021robustnet}                               & 72.10                       \\ \hline
\multicolumn{1}{|l|}{MRFP (Ours)}  & \multicolumn{1}{l|}{74.28} \\ \hline
\multicolumn{1}{|l|}{MRFP+ (Ours)} & \multicolumn{1}{l|}{74.85} \\ \hline
\end{tabular}
\caption{Source-domain performance of models trained on the GTAV dataset with Resnet-50 backbone .}
\label{table:SourceG}
\end{table}

\begin{table}[ht]
\centering
\begin{tabular}{|c|c|}
\hline
Models (S)                         & S                          \\ \hline
Baseline                           & 78.37                      \\ \hline
ISW \cite{choi2021robustnet}                               & 77.48                      \\ \hline
\multicolumn{1}{|l|}{MRFP (Ours)}  & \multicolumn{1}{l|}{75.87} \\ \hline
\multicolumn{1}{|l|}{MRFP+ (Ours)} & \multicolumn{1}{l|}{74.05} \\ \hline
\end{tabular}
\caption{Source-domain performance of models trained on the Synthia dataset with Resnet-50 backbone .}
\label{table:SourceS}
\end{table}
\vspace{-4mm}
\begin{table}[ht]
\centering
\begin{adjustbox}{max width=0.48\textwidth}
\begin{tabular}{l|l|l|l|l|l}
\toprule
Model(GTAV) & B              & C              & S              & M              & Avg            \\ \midrule
U-Net \cite{ronneberger2015u}       & 25.32          & 28.49          & 19.48          & 30.80          & 26.02          \\ \midrule
U-Net with MRFP  & \textbf{28.87} & \textbf{28.79} & \textbf{19.60} & \textbf{36.01} & \textbf{28.31} \\ \bottomrule
\end{tabular}
\end{adjustbox}
\caption{Performance comparison of domain generalization methods in terms of mIoU (\%) using ResNet-50 backbone. Models are trained on G $\rightarrow$ \{B, C, S, M \}.}
\label{table:Unet}
\end{table}

\subsection{Qualitative Results}

\begin{figure*}[ht]
    \centering
    \includegraphics[scale=0.48]{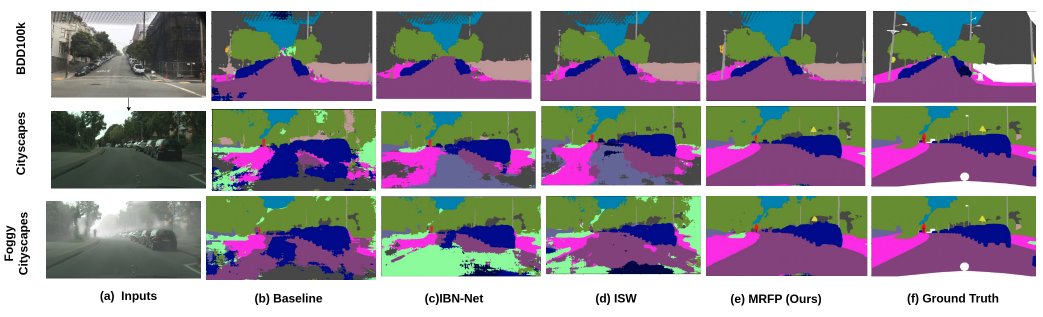}
    \caption{Qualitative comparison with different domain generalization methods, trained on GTAV \cite{richter2016playing} with the backbone as ResNet-50, tested on BDD-100K \cite{yu2020bdd100k}, Cityscapes \cite{cordts2016cityscapes} and Foggy Cityscapes \cite{sakaridis2018semantic}. MRFP consistently generates superior predictions compared to other methods, especially on Cityscapes and Foggy Cityscapes.}
    \label{fig:seg_pred}
\end{figure*}

Figure \ref{fig:seg_pred} shows the qualitative results of MRFP benchmarked against other state-of-the-art domain generalization methods. The results indicate that MRFP outperforms other methods in unseen real-world datasets, especially in adverse weather conditions as seen in Foggy Cityscapes.

\subsection{Ablation study}

\textbf{Choice of network design:} In designing the HRFP module, we chose a four layer encoder-decoder structure guided by GradCAM visualizations. Emperically, incorporating 4+ layers mostly captures redundant fine features leading to performance saturation with only a marginal +0.3\% OOD performance gain, albeit with increased GPU memory usage during training. Conversely, 2 or 3 layers result in -1.5\% OOD performance. 
In the HRFP module, the number of channels in each layer remains consistent, except for the layers added to the base network, ensuring compatibility with the summation operation.

Table \ref{table:GTAtableablation} shows the relative performance of three different up-scaling factors for the bilinear interpolation of the (High Resolution Feature Perturbation) HRFP block. The proposed method HRFP is empirically set to have a overall scale increase of 2 in its latent space. In other words, the spatial resolution, over the course of the first four encoder layers, is ultimately increased to double the spatial resolution of the input to the HRFP block. This can be termed as overall scale-factor increase. To evaluate the memory consumption to accuracy tradeoffs, we present additional results on overall scale factors of 1.5 and 2.5 respectively in Table \ref{table:GTAtableablation}. It is observed that while a higher overall scale-factor of 2.5 provides a slight improvement in the OOD performance, the memory requirement during training is substantially higher. 
%We anticipate that this performance increase is due to the increase in the scale factor, which aids the network in perturbing even higher domain-variant frequencies. In contrast,
An up-scaling factor of 1.5 provides a slight drop in the OOD performance while the memory requirements tend to be on the lower end.
%likely due to its inability to focus on domain-variant high-frequency features. 
Overall, the sensitivity to the up-scaling factor seems to be minimal, but in the proposed method we adopt an overall scale factor of 2 as it helps us decrease the memory usage compared to 2.5 and add the HRFP+ decoder perturbation which yields significant OOD performance improvements.

Table \ref{table:NoiseAblation} shows the benefit of using the proposed MRFP/MRFP+ for domain generalization. In Table \ref{table:NoiseAblation}, L-MRFP is learnable MRFP which refers to the HRFP block of MRFP being learnable and not randomized every iteration. RGN refers to random Gaussian noise being added to features as feature perturbations instead of the proposed MRFP/MRFP+ module to the baseline model (DeepLabV3+).
It is seen that feature perturbation is helpful for robustness to domain shifts as L-MRFP performs inferior to the proposed RGN and the proposed MRFP/MRFP+. MRFP/MRFP+ out-of-domain(OOD) performance surpasses L-MRFP and RGN by approximately 4\%. This suggests that adding random noise to the features as perturbations does increase the OOD performance but has the issue of redundancy, and distorting the semantics of the domain-invariant features which can negatively impact OOD performance thus leaving room for improvement. To fill this gap, the proposed MRFP/MRFP+ technique predominantly perturbs domain-specific features while minimizing the semantic distortion that can occur to domain-variant features by random perturbtion.\\
The proposed HRFP block and the NP+ style perturbation technique is introduced in the initial layers as they predominantly contain style information and fine-grained domain-specific features. To restrict these domain-specific features from propagating and influencing the later layers of the network, we add the HRFP/HRFP+ block with style perturbation techniques at the initial layers. This setting was empirically found to be the best.
\vspace{-4mm}
\begin{table}[h]
\centering
\begin{adjustbox}{max width=0.48\textwidth}
\begin{tabular}{l|l|l|l|l}
\toprule
Models(GTAV) &  B     & C     & M        & Avg   \\ \midrule

MRFP (OSF=1.5)   & 38.29	 & 40.27 & 41.95  & 40.17 \\ \midrule
MRFP (OSF=2.5)   & 39.21  & 39.84 & 43.20 & 40.75 \\ \midrule

MRFP   &  38.80 & 40.25 & 41.96  & 40.33 \\ \midrule
MRFP+ &    \textbf{39.55} & \textbf{42.40} & \textbf{44.93}   & \textbf{42.29}         \\ \bottomrule
\end{tabular}
\end{adjustbox}
\caption{Performance comparison of domain generalization methods in terms of mIoU using ResNet-50 backbone. Models are trained on G $\rightarrow$ \{B, C, M\}. OSF= Overall scale-factor increase.}
 \label{table:GTAtableablation}
\end{table}

\begin{table}[ht]
\centering
\begin{tabular}{l|l|l|l|l|l}
\toprule
Model(GTAV)                          & B              & C              & S              & M              & Avg            \\ \midrule
Baseline                             & 31.44          & 34.66          & 25.84          & 32.93          & 31.21          \\ \midrule
L-MRFP                       & 33.38          & 38.21          & 25.40          & 38.04          & 33.75          \\ \midrule
RGN     & 34.70          & 36.84          & 25.36          & 41.81          & 34.67          \\ \midrule
MRFP(Ours)                           & 38.80          & 40.25          & 27.37          & 41.96          & 37.09          \\ \midrule
MRFP+(Ours)                          & \textbf{39.55} & \textbf{42.20} & \textbf{30.22} & \textbf{44.93} & \textbf{39.27} \\ \bottomrule
\end{tabular}
\caption{Performance comparison of domain generalization methods in terms of mIoU using ResNet-50 backbone. Models are trained on G $\rightarrow$ \{B, C, S, M \}. RGN=Random Gaussian Noise, L-MRFP=Learnable MRFP.}
 \label{table:NoiseAblation}
\end{table}

\subsection{Computational Complexity}

Table \ref{table:Computation} shows the computational complexity comparison of the proposed MRFP technique with other domain generalization methods. 
Since MRFP is only used for training, and only the baseline segmentation model DeepLabV3+ is used for inference, the number of parameters and the computational cost (GMACs) remains the same as the baseline model. Although the parameters of SAN-SAW \cite{peng_semantic-aware_2022} is considerably less compared to MRFP and the baseline model DeepLabV3+ model, its computational cost and training time is extremely high. 

\begin{table}[h]
\centering
\begin{tabular}{|l|l|l|}
\hline
Models   & \# of Params & GMACs   \\ \hline
Baseline & 40.35M       & 554.31 \\ \hline
IBN-Net \cite{pan2018two}  & 45.08M       & 555.64 \\ \hline
ISW \cite{choi2021robustnet}     & 45.08M       & 555.56 \\ \hline
SAN-SAW \cite{peng_semantic-aware_2022} & 25.63M       & 843.72 \\ \hline
WildNet \cite{lee2022wildnet} & 45.21M       & 554.32 \\ \hline
MRFP     & 40.35M       & 554.31 \\ \hline
\end{tabular}
\caption{Inference computation cost comparisons of MRFP and other contemporary methods. }
\label{table:Computation}
\end{table}

%----------------------------------------------------------------------
\subsection{Comparison With Alternative Domain Generalization Techniques}

The domain generalization performance of the proposed MRFP/MRFP+ technique is compared with numerous state-of-the-art DG methods.
The augmentations used for training and re-implementation are restricted to standard augmentations, and all settings are consistent with ISW \cite{choi2021robustnet}.  We do not perform a performance comparison with ProRandConv \cite{choi2023progressive} on the Foggy Cityscapes dataset \cite{sakaridis2018semantic} because its open-source code is not available. We also do not conduct a performance comparison between MRFP and \cite{huang2023style}, because of mainly two reasons: (1) in addition to using data augmentations used in \cite{choi2021robustnet}, extra strong style augmentations are used to enhance the style information of urban-scene images, using Automold road augmentation library\footnote{Automold Road augmentation library : \url{https://github.com/UjjwalSaxena/Automold--Road-Augmentation-Library}}, and (2) open source code is not available for re-implementation without the extra strong style augmentations used. 
In the future, we intend to use these augmentations for style variations on source domains. 
%%%%%%%%% REFERENCES
%{\small
%\bibliographystyle{ieeenat_fullname}
%\bibliography{main}
%}

%\end{document}

\end{document}